\documentclass{article}

\usepackage{microtype}
\usepackage{graphicx}
\usepackage{subfigure}
\usepackage{booktabs} 
\usepackage{tcolorbox}
\usepackage{multirow}

\usepackage{hyperref}

\usepackage[accepted]{icml2025}

\usepackage{amsmath}
\usepackage{amssymb}
\usepackage{mathtools}
\usepackage{amsthm}

\usepackage[capitalize,noabbrev]{cleveref}

\theoremstyle{plain}
\newtheorem{theorem}{Theorem}[section]

\newtheorem{lemma}[theorem]{Lemma}
\newtheorem{corollary}[theorem]{Corollary}
\theoremstyle{definition}

\theoremstyle{remark}

\usepackage[textsize=tiny]{todonotes}

\icmltitlerunning{MCTS-EP: Empowering Embodied Planning with Online Preference Optimization}

\newcommand{\argmin}{\mathop{\mathrm{argmin}}}
\newcommand{\argmax}{\mathop{\mathrm{argmax}}}

\makeatletter
\def\ICML@appearing{}
\makeatother
\begin{document}

\twocolumn[
\icmltitle{MCTS-EP: Empowering Embodied Planning with Online Preference Optimization}



\icmlsetsymbol{equal}{*}
\icmlsetsymbol{project lead}{+}
\icmlsetsymbol{research lead}{‡}

\begin{icmlauthorlist}
\icmlauthor{Hang Xu}{equal,fdu,agibot}
\icmlauthor{Zang Yu}{equal,agibot}
\icmlauthor{Yehui Tang}{project lead,agibot}
\icmlauthor{Pengbo Hu}{agibot}
\icmlauthor{Yuhao Tang}{research lead,independent}
\icmlauthor{Hao Dong}{cfcs,PKU-AgiBot}
\end{icmlauthorlist}

\icmlaffiliation{fdu}{School of Management, Fudan University, Shanghai, China}
\icmlaffiliation{agibot}{AgiBot, Shanghai, China}
\icmlaffiliation{independent}{Independent Researcher}
\icmlaffiliation{cfcs}{CFCS, School of Computer Science, Peking University, Beijing, China}
\icmlaffiliation{PKU-AgiBot}{PKU-AgiBot Lab, Peking University, Beijing, China}

\icmlcorrespondingauthor{Hao Dong}{hao.dong@pku.edu.cn}
\icmlkeywords{Embodied Agent, MCTS, Direct Preference Optimization, Multi-Modal Planning}

\vskip 0.3in
]
\printAffiliationsAndNotice{\icmlEqualContribution} 



\begin{abstract}
This paper introduces MCTS-EP, an online learning framework that combines large language models (LLM) with Monte Carlo Tree Search (MCTS) for training embodied agents. MCTS-EP integrates three key components: MCTS-guided exploration for preference data collection, efficient multi-modal reasoning mechanism, and iterative training pipeline based on preference optimization. We theoretically prove that MCTS-EP achieves better performance bounds than conventional on-policy algorithms when the loss function is strongly convex, and demonstrate that it can be formulated as a search-enhanced variant of GAIL. 
MCTS-EP achieves state-of-the-art performace across serval benchmarks. In ALFWorld, it achieves 92\% and 87\% success rates for textual and visual tasks. In WebShop, it reaches an average reward of 0.81. MTCS-EP also reduces average interaction steps from from 18.7/19.5 to 10.2/9.9 steps in textual/visual ALFWorld.\footnote{Code available at: \url{https://github.com/xuhang-2/Embodied-Agent-Planning}}
\end{abstract}

\section{Introduction}
Recently, large language models (LLM) trained on internet-scale data have demonstrated strong capabilities in commonsense reasoning\cite{touvron2023llama, bai2023qwen, team2024gemma, wei2022chain}. Furthermore, vision-language models (VLM) effectively bridge the gap between language instructions and visual observations, enabling more robust multi-modal reasoning \cite{wang2024qwen2, li2023blip, liu2024visual, zhu2023minigpt}. Building on these advancements, embodied agents can now perform complex decision-making tasks in interactive environments based on general commonsense knowledge \cite{wu2023tidybot, li2024llava, stella2023can}. 

A key challenge for decision-making tasks is that agents must reason over multi-modal information and sequentially decompose tasks into actionable steps \cite{jansen2020visually, huang2022language, li2022pre}. In the domain of text generation, the state space of token is exponentially larger\cite{guo2025deepseek}. Employing Monte Carlo Tree Search (MCTS) faces substantial computational cost. However, embodied agent's action space is discrete and small when receive environment feedback. This making MCTS suitable for exploration in environments. Recent works have explored using MCTS \cite{coulom2006efficient, kocsis2006bandit} to enhance LLM reasoning\cite{zelikman2024quiet, zhao2024large, zhang2024rest}. But these methods are not exclusively focused on embodied scenarios.

Several approaches have emerged to enhance embodied agents through feedback, preference optimization, and cross-modal learning. Reflexion \cite{shinn2024reflexion} uses linguistic feedback and episodic memory to enable agents to learn from trial-and-error without expensive model fine-tuning. IPR \cite{xiong2024watch} introduces step-level rewards and preference optimization to provide detailed supervision instead of outcome-based signals. SayCan \cite{ahn2022can} uses visual navigation to collect information in the house to generate grounded plans. EMMA \cite{yang2024embodied} takes a cross-modality approach by distilling LLM's knowledge from text world to guide visual world learning. 

Despite advances in embodied planning algorithms, several limitations remain.
First, generating high-quality data is a persistent challenge. Most methods rely on imitation learning \cite{hussein2017imitation}, which depends entirely on expert demonstrations. Imitation learning faces quadratic error accumulation challenge in long-horizon tasks \cite{ross2010efficient}. Besides, these approaches lack sufficient exploration of interactive environments and fail to leverage knowledges from trial-and-error \cite{song2024trial}. 
Second, many existing works focus on TextWorld \cite{cote2019textworld}, where agents depend on structured textual feedback and predefined candidate actions to guide planning. However,  visual agents must infer actions and environmental feedback directly from camera inputs in real-world scenarios. EMMA \cite{yang2024embodied} partially addresses this limitation through a multi-modal approach but struggles with repeated trial-and-error attempts, which are impractical in real-world applications.
Lastly, methods like IPR leverage direct preference optimization (DPO) \cite{rafailov2024direct} for preference learning. But they rely on static datasets or random sampling to collect preference pairs, limiting the model to passively learn from predefined trajectories \cite{xiong2024watch}. This results in offline trainings which can fail with high probability \cite{xie2024monte}. In contrast, an online framework enables iterative learning through self-exploration, enhancing planning capabilities and generalization.

To address these limitations, we propose a multi-modal framework (MCTS-EP) that generates preference data and trains LLM for embodied agents online.  
MCTS-EP includes three components: 
1) Preference Data Collection based on MCTS: 
We employ iteratively training LLM as policy model to guide MCTS to generate preference action pairs and success trajectories. We make a series of improvements to MCTS to improve the efficiency of data collection, 
2) Selective State Representation for Multi-Modal Reasoning: We propose a new multi-modal planning pipeline that balances reasoning speed with memory utilization. 
3) MCTS-Enhanced DPO: We design an MCTS-Enhanced DPO framework that combines expert fine-tuning, preference data collection, and iterative optimization to improve model's planning capabilities.

The main contributions of this work can be summarized as:
\begin{enumerate}
    \item Preference Data Collection based on MCTS: We propose an efficient and robust preference data collection method for embodied agent. In the process, MCTS can be seen as an implicit reward model.
    \item MCTS-Enhanced DPO: We design a pipeline that maintains inference efficiency while incorporating memory during reasoning. The training process approximates offline DPO as multi-round online training, improving model performance in both accuracy and task efficiency.
    \item We evaluate MCTS-EP on the ALFWorld and WebShop \cite{shridhar2020alfworld,yao2022webshop}. Experimental results show that MCTS-EP achieves state-of-the-art performance in both score, success rate and task efficiency.
\end{enumerate}

\section{Related Work}
\textbf{Long-horizon Reasoning with Language Models.}
Recent LLMs have significantly advanced long-horizon reasoning  \cite{touvron2023llama, bai2023qwen}. Chain-of-thought prompting \cite{wei2022chain} enables step-by-step task decomposition. Several works combine MCTS with LLM for systematic exploration \cite{zelikman2024quiet, zhao2024large, zhang2024rest}, while \cite{jiao2024learning} proposes a planning-based framework that synthesizes process rewards from offline trajectories to enhance reasoning without costly human annotations. Reflexion \cite{shinn2024reflexion} uses feedback and memory for iterative improvement. However, these methods primarily target language-only tasks. VLM \cite{wang2024qwen2, li2023blip} bridge visual and linguistic understanding, enabling multi-modal reasoning \cite{chen2024spatialvlm,chen2024large}.

\noindent \textbf{Task Planning for Embodied Agents.}
Task planning for embodied agents requires decomposing high-level goals into executable action sequences while considering environmental constraints. Traditional approaches rely heavily on structured action spaces and expert demonstrations \cite{hussein2017imitation}, limiting their applicability in dynamic environments. Recent works like SayCan \cite{ahn2022can} have made progress incorporating visual grounding for more robust planning, while IPR \cite{xiong2024watch} introduces preference optimization to provide fine-grained supervision. EEMA \cite{yang2024embodied} takes a cross-modal approach to transfer knowledge from text to visual domains, though it often requires multiple attempts to complete tasks successfully. APRICOT \cite{wang2024apricot} further advances this line of work by integrating active preference learning with constraint-aware planning, enabling robots to infer user preferences from sparse demonstrations while ensuring feasibility in real-world environments.

\section{Method}

\begin{figure*}[ht]    
    \centering
    \includegraphics[width=\textwidth]{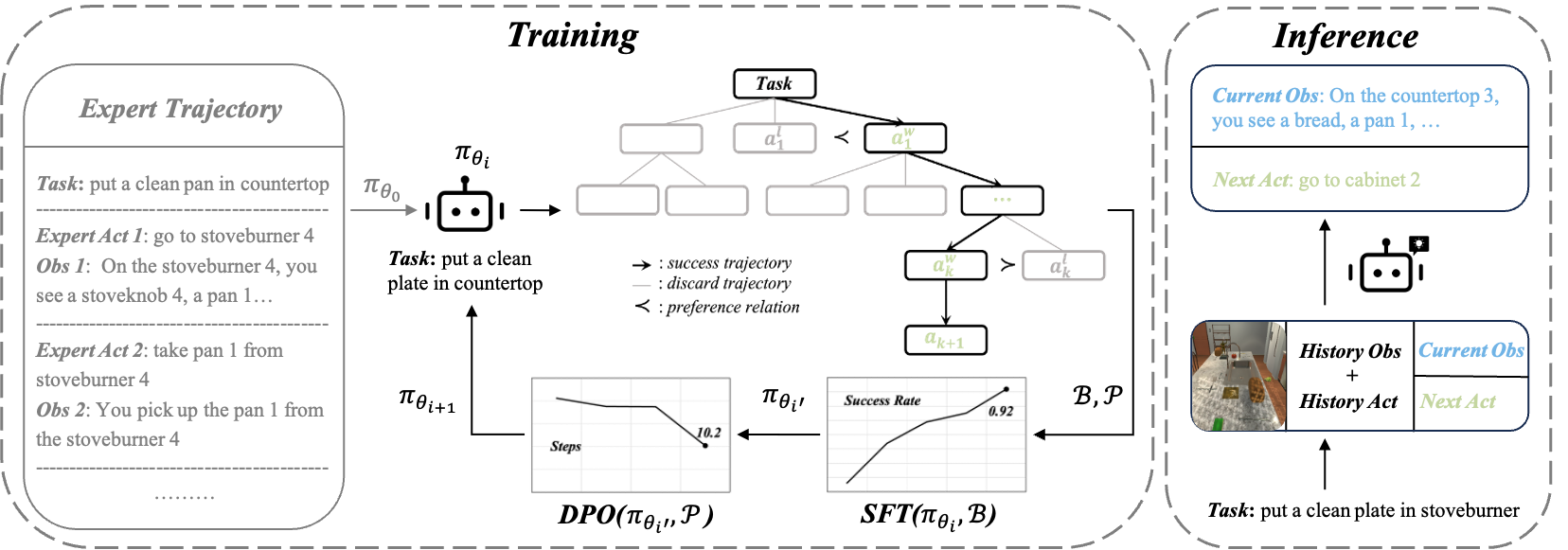}
    \caption{\textbf{Overview framework}. The framework consists of three key components: (1) MCTS-guided preference data collection, (2) selective state representation for multi-modal reasoning, and (3) iterative training pipeline.}
    \label{fig:example}
\end{figure*}

\subsection{Preliminary}
In this section, we introduce the basic settings of MCTS-EP. Embodied task planning can be formulated as a partially observable Markov decision process (POMDP), defined by a tuple $\langle \mathcal{U}, \mathcal{A}, \mathcal{O}, \mathcal{S}, \mathcal{T}, \mathcal{R} \rangle$. $\mathcal{U}$ is instruction space. $\mathcal{A}$ is action space. $\mathcal{O}$ is observation space. $\mathcal{S}$ is state space. $s_t \in \mathcal{S}$ can be written as:
\begin{equation}
s_t=(u, a_1, o_1, ..., a_t, o_t),
\end{equation}
where $u_t \in \mathcal{U}$ is a natural language instruction, an example is "put a clean pan in countertop", $a_t \in \mathcal{A}$ is action in step t, $o_t \in \mathcal{O}$ is observation in step t, which can be either textual or visual.

$\mathcal{T}: S \times A \times S \rightarrow [0,1]$ is the transition function. $\mathcal{R}: S  \rightarrow [0,1]$ is the reward function. In this paper, we use outcome reward function, only applying to terminal states. We define initial policy model ${\pi}_{\theta_0}$ and iterative policy model ${\pi}_{\theta_i}$ as LLM or VLM. The objective function is
\begin{equation}
\min_{\theta_i} \mathbb{E}_{s_t \sim \mathcal{S}} [D_{KL}(\pi_{\theta_i}(a^{\ast}_t|s_t) \| \pi^*(a^{\ast}_t|s_t))],
\end{equation}
where $\pi_{\theta_i}(a^{\ast}_t|s_t)$ is the predicted action distribution, $\pi^{\ast}(a^{\ast}_t|s_t)$ is the optimal policy distribution, $a^{\ast}_t$ is the ground truth action.

\subsection{Preference Data Collection based on MCTS}
\label{subsec:data_generation}
To generate step-level preference data, we decompose the planning process into several phases in the React-style \cite{yao2022react}. In every planning process, we sample candidate action from the distribution of ${\pi}_{\theta_i}$. We employ MCTS to collect preference data, capturing long-term action consequences through value propagation.
\subsubsection{Search Tree Construction}

Starting from initial state node $s_0$, MCTS builds a search tree through iterative four phases search: selection, expansion, simulation, and backup.

\textbf{Selection}: Starting from the root node, the algorithm recursively selects child nodes until reaching a leaf node using Predictor + Upper Confidence bounds applied to Trees (PUCT) \cite{rosin2011multi}: 
$$
a_t^{\ast} = \argmax\limits_{a \in \mathcal{A}} \left[Q(s_t,a) + c_{puct}\pi_{{\theta}_{i}}(a|s_t)\sqrt{\frac{N(s_t)}{1 + N(s_t,a)}}\right],
$$
where $Q(s_t,a)$ is estimated action value, 
$N(s_t)$ is visit count of state $s_t$,
$N(s_t,a)$ is visit count of action $a$ at state $s_t$, 
$c_{puct}$ is coefficient, 
$\pi_{\theta}(\cdot|s_t)$ is action distribution given $s_t$.

\textbf{Expansion}: We introduce a self-critic \cite{zhang2024rest} to improve expansion efficiency. For each leaf node $s_t$ during selection, we employ the critic LLMs $\pi_{{\theta}_{c}}$ to evaluate whether expansion is necessary:
\begin{equation}
  \text{expand}(s_t) = \begin{cases}
\text{true} & \text{if } \pi_{{\theta}_{c}}(s_t) > \tau \\
\text{false} & \text{otherwise}, 
\end{cases}  
\end{equation}
\noindent where $\tau$ is a threshold. If $\text{expand}(s_t)$, the algorithm proceeds with standard expansion by sampling actions from policy model $\pi_{\theta}(\cdot|s_t)$ and adding corresponding child nodes.

\textbf{Simulation}: We employ greedy search to guide simulation. This approach combines the reasoning of LLM with the exploration of MCTS. The simulation continues until either reaching a terminal state $s_{\text{terminal}}$ or the maximum depth $d_{max}$. The reward function $r_o$ evaluates the final state to provide a reward:
\begin{equation}
r_o(s_{\text{terminal}}) = \begin{cases}
1 & \text{if task completed} \\
0.5 & \text{if task partially completed} \\
0 & \text{if task not completed}.
\end{cases} 
\end{equation}

\textbf{Backup}: After simulation, reward are propagated back through action path. For each node $s_t$ in the backup path, we update its statistics as follows:
\begin{equation}
\begin{aligned}
N(s_t) &\leftarrow N(s_t) + 1 , \\
V(s_t) &\leftarrow V(s_t) + \gamma \cdot r(s_{t+1}) , \\
Q(s_t, a_t) &\leftarrow Q(s_t, a_t) \\&+ \frac{1}{N(s_t, a_t)}[V(s_{t+1}) - Q(s_t, a_t)],
\end{aligned}
\end{equation}
where $V(s_{t+1})$ is the simulated value from the child node, $r(s_{t+1})$ is discount reward propagated back through the simulation path with discount factor $\gamma$.

\subsubsection{Data Collection from Search Tree}
We can extract two datasets from constructed search tree.
\textbf{Success Trajectory Dataset $\mathcal{B}$}: 
For a single natural language task, multiple valid solution paths may exist. For instance, given the task "put a clean pan on the table", there could be multiple pans in the environment. While an expert trajectory might demonstrate picking up a pan from the countertop, the agent could alternatively retrieve it from other valid containers.

We collect all successful paths discovered during Monte Carlo tree search into a success trajectory buffer $\mathcal{B} = \{(u, a_1, ..., o_T)^i\}_{i=1}^N$, where $N$ is the number of successful trajectories. These high-quality trajectories serve as online training data and can be used to fine-tune iterative policy network $\pi_{\theta_{i}}$.

\textbf{Preference Pairs Dataset $\mathcal{P}$}: 
At any given state, multiple actions may lead to successful task completion. During MCTS's backup, each node along the path receives updated reward estimates. We leverage this information to create preference pairs at branching nodes. Specifically, for each decision point $s_t$, we can generate ordered pairs $(a_i, a_j)$ where actions are ranked according to their estimated action values

\subsection{Selective State Representation for Multi-Modal Reasoning}
In real-world applications, robots get observation from continuous camera feeds. At each timestep \(t\), the agent receives a raw image observation \(o_t\). Policy model \(\pi_{\theta_i}\) processes \(o_t\) to predict next action \(a_t\) and extract key textual information \(\tau_t\) (like object labels or spatial relationships):
\begin{equation}
a_t, \tau_t = \pi_\theta(s_t)
\end{equation}

For long-horizon tasks, storing all historical observations for reasoning is computationally expensive. We optimize computational cost by storing only textual descriptions \(\{\tau_1, ..., \tau_{t-1}\}\) and actions from previous steps, while keeping the full image observation \(o_t\) only for current timestep
The state representation evolves as follows:
\begin{itemize}
\item Previous state: \(s_{t-1} = (u, a_1, \tau_1, ..., a_{t-1}, \tau_{t-1})\)
\item Current state: \(s_t = s_{t-1} \cup \{a_t, o_t\}\)
\end{itemize}

Visual-language models can thus operate efficiently while preserving critical environmental details.

\subsection{MCTS-Enhanced DPO}
MCTS-EP is designed to iteratively improve agent's performance by MCTS for exploration and data generation. Staring from an initial policy model $\pi_{\theta_0}$, the framework alternates between data generation and model updates. At iteration \(t\), MCTS-EP performs following steps:

\begin{enumerate}
    \item Data generation with MCTS (Section~\ref{subsec:data_generation}): Using current policy model $\pi_{\theta_{i}}$, MCTS is applied to generate success trajectory data $\mathcal{B}$ and preference pairs data $\mathcal{P}$.
    \item Imitation learning on expert trajectories: The training process warms up by leveraging expert trajectories $\mathcal{D}_\text{expert}$ for pre-trained LLM/VLM to initialize policy model $\pi_{\theta_{0}}$.
    \item Fine-tuning on success trajectory data $\mathcal{B}$: Success trajectory data is used to fine-tune the policy model \(\pi_{\theta_i}\), incorporating successful attempt to enhance model performance through iterative updates.
    \item Preference optimization on preference data $\mathcal{P}$: Preference pairs data \(\mathcal{P}_t\) is used to further fine-tune the model, aligning the policy with task-specific preferences. 
\end{enumerate}

\subsubsection{Imitation Learning with Expert Trajectory}
The first stage of the process involves Supervised Fine-Tuning (SFT) based on expert demonstrations $\mathcal{D}_\text{expert}$. In embodied scenarios, gathering large quantities of high-quality demonstrations is often costly and challenging. Therefore, small datasets serve to train the initial policy model $\pi_{\theta_{0}}$ to produce properly formatted, task-compliant outputs. The performance of policy model $\pi_{\theta_{0}}$ relies on following continuous improvement process.

Specifically, we denote expert trajectory $e_u \in \mathcal{D}_\text{expert}$ as $e_u = (u, a_1^{\ast}, a_2^{\ast},... , a_k^{\ast})$, where $a_t^{\ast}$ is the best action at step t. Based on $e_u$, we can get expert terminal state $s_{u_t} = (u, a_1^{\ast}, o_1, ..., a_k^{\ast}, o_k)$. Our object is to minimize loss function:
\begin{align}
\mathcal{L}_{\text{SFT}}(\theta) &= -\mathbb{E}_{e_u \sim \mathcal{D}_\text{expert}} [\log \pi_{\theta}(e_u|u)] \nonumber \\
&= -\mathbb{E}_{e_u \sim \mathcal{D}_\text{expert}} \left[\sum_k \log \pi_{\theta}(a_k|s_{u_{k-1}})\right],
\end{align}
where $s_{u_{k-1}} = (u, a_1^{\ast}, o_1, ..., a_{k-1}^{\ast}, o_{k-1})$. SFT provide a good starting point for further optimization and exploration.

\subsubsection{Fine-Tuning with Success Trajectory Data}
After obtaining the initial policy model \(\pi_{\theta_0}\), MCTS leverages the initial policy \(\pi_{\theta_0}\) to guide the search process and generate trajectories. We can get preference pairs dataset $\mathcal{P}$ and success trajectory dataset $\mathcal{B}$. Each success trajectory in $\mathcal{B}$ can be seen as a terminal state \(s_T\) when the trajectory satisfies task-specific objectives. \(s_T\) is composed of a sequence of observation-action pairs. 

Formally, the objective is to minimize the negative log-likelihood:
\begin{equation}
\mathcal{L}_\text{SFT}(\theta) = -\mathbb{E}_{s_T \sim \mathcal{B}} \left[ \sum_{t=1}^T \log \pi_\theta(a_t | s_{t-1}) \right].
\end{equation}

Compared to the initial imitation learning stage, fine-tuning with success trajectories enables the policy to adapt based on its own exploration of the environment. The use of MCTS allows for the discovery of trajectories that may not be present in the expert dataset \(\mathcal{D}_\text{expert}\), enabling the policy to generalize beyond the constraints of the expert demonstrations. Besides, since the outcome reward model $r_o$ directly evaluates whether a trajectory achieves the task objective, the fine-tuning process optimizes the policy with respect to task-specific goals rather than purely imitating expert behavior.

\subsubsection{Preference Optimization with Preference Data}
DPO improves the policy model's commonsense understanding while enhancing task efficiency, such as reducing steps needed for object finding.

To achieve this alignment, preference data \(\mathcal{P}\) is utilized. Each preference in \(\mathcal{P}\) is represented as a tuple \((a_t^w, a_t^l, s_{t-1})\), where \(a_t^w\) denotes the winning action that leads to a higher reward,  \(a_t^l\) denotes the losing action with a lower reward.  The preference data is used to construct a contrastive objective that encourages the policy to assign higher probabilities to \(a_t^w\) over \(a_t^l\) in the same context \(s_{t-1}\). Formally, the DPO objective is defined as:  
\begin{equation}
\begin{split}
\mathcal{L}_\text{DPO}(\theta) = -\mathbb{E}_{(a_t^w, a_t^l, s_{t-1}) \sim \mathcal{P}} [
\log(\sigma(\beta \cdot \\
(\log(\frac{\pi_\theta(a_t^w|s_{t-1})}{\pi_{ref}(a_t^w|s_{t-1})}) - 
\log(\frac{\pi_\theta(a_t^l|s_{t-1})}{\pi_{ref}(a_t^l|s_{t-1})}))))
], 
\end{split}
\end{equation}
where \(\pi_\theta(a_t^w | s_{t-1})\) and \(\pi_\theta(a_t^l | s_{t-1})\) are the conditional probabilities assigned by the policy \(\pi_\theta\) to $a_t^w$ and $a_t^l$, respectively.

The preference data \(\mathcal{P}\) is typically derived from policy model exploration or expert demonstrations. For a given state \(s_{t-1}\), the model evaluates multiple candidate actions and their resulting outcomes. A preference is constructed whenever \(Q(s_t, a_t^w) > Q(s_t, a_t^l)\), indicating that \(a_t^w\) leads to a more favorable outcome compared to \(a_t^l\).

\begin{algorithm}[tb]
   \caption{MCTS-Enhanced DPO}
   \label{alg:MCTS-EP}
\begin{algorithmic}
   \STATE {\bfseries Input:} Initial policy model $\pi_{\theta_0}$, expert demonstrations $\mathcal{D}_\text{expert}$
   \STATE {\bfseries Output:} Optimized policy model $\pi_{\theta}$
   
   \STATE {\bfseries Phase 1: Imitation Learning}
   \FOR{$e_u=(u,a_1^*,...,a_k^*) \in \mathcal{D}_\text{expert}$}
       \STATE $s_{u_{k-1}} \gets (u,a_1^*,o_1,...,a_{k-1}^*,o_{k-1})$ 
       \STATE Minimize $\mathcal{L}_\text{SFT}(\theta) = -\sum_k \log \pi_{\theta}(a_k|s_{u_{k-1}})$
   \ENDFOR

   \STATE {\bfseries Phase 2: Iterative Improvement}
   \FOR{iteration $i=1$ to $T$}
       \STATE $\mathcal{B}, \mathcal{P} \gets \text{MCTS}(\pi_{\theta_i})$ \COMMENT{Step 1: Generate success and preference data}
       
       \FOR{$s_T \in \mathcal{B}$}
           \STATE Minimize $\mathcal{L}_\text{SFT}(\theta) = -\sum_{t=1}^T \log \pi_\theta(a_t|s_{t-1})$
           \COMMENT{Step 2: Fine-tune with Success Trajectories}
       \ENDFOR
       \FOR{$(a_t^w, a_t^l, s_{t-1}) \in \mathcal{P}$}
           \STATE $p_w \gets \log(\pi_\theta(a_t^w|s_{t-1})/\pi_\text{ref}(a_t^w|s_{t-1}))$
           \STATE $p_l \gets \log(\pi_\theta(a_t^l|s_{t-1})/\pi_\text{ref}(a_t^l|s_{t-1}))$
           \STATE Minimize $\mathcal{L}_\text{DPO}(\theta) = -\log(\sigma(\beta(p_w - p_l)))$ \COMMENT{Step 3: Direct Preference Optimization}
       \ENDFOR
       
       \STATE $\theta_{i+1} \gets \theta_i$ \COMMENT{Update model parameters}
   \ENDFOR
    \STATE \textbf{return} $\pi_{\theta}$
\end{algorithmic}
\end{algorithm}

Complete MCTS-EP algorithm is summarized in \cref{alg:MCTS-EP}. The algorithm combines imitation learning, fine-tuning with success trajectories, and direct performance optimization to refine the policy model iteratively. MCTS-guided exploration enables the model to learn from successful experiences. Preference optimization helps the model understand and incorporate task-specific preferences. This integrated approach addresses the challenges of both exploration and optimization in complex embodied environments.

\section{Theory Analysis}
\subsection{MCTS Performance Bounds}
We provide theoretical performance guarantees for MCTS-based policies compared to conventional on-policy algorithms. Our analysis demonstrates that MCTS exhibits superior bounds under certain conditions.

\begin{theorem}
\label{theorem:Performance}
Let $\pi_{\theta}$ be a policy represented by a language model and $\pi_{mcts}$ be the policy derived from Monte Carlo Tree Search with Upper Confidence Bounds. For any state $s$, if the loss function $l$ is strongly convex, then:
$$\mathbb{E}[l(s, \pi_{mcts})] \leq \mathbb{E}[l(s, \pi_{\theta})].$$
\end{theorem}

The complete proof of \cref{theorem:Performance} is presented in \cref{app:Performance}. Our proof follows a similar structure to DAgger \cite{ross2011reduction}, but extends it to the MCTS setting. This theoretical guarantee provides a formal foundation for the empirical success of MCTS in various decision-making scenarios and offers insights into why MCTS-enhanced policies tend to outperform standard policy networks.

\subsection{MCTS-EP as Search-Enhanced GAIL Variant}
Generative Adversarial Imitation Learning (GAIL) \cite{ho2016generative} optimizes policies via adversarial training to match the state-action occupancy measure of expert demonstrations. Traditional GAIL uses a discriminator to distinguish between expert and learned policy trajectories. In this section, we show that MCTS-EP can be viewed as a search-enhanced variant of GAIL, where the MCTS serves as an implicit discriminator through its value estimation process. This connection is formally established in the following theorem:
\begin{theorem}
\label{theorem:gail}
Let cost regularization is $$\psi(a) = c*\frac{\sqrt{ln t}}{t*\rho_{\pi}(s,a)} + \mathbb{E}_{\pi_E}[\log \pi_{E}(a|s)],$$ MCTS-EP can be approximately formulated as GAIL framework.
\end{theorem}
The complete proof can be found in \cref{app:gail}. This formulation reveals a two-stage optimization process. The Lower Confidence Bound LCB \cite{rashidinejad2021bridging} policy first generates a value function that serves as the cost function in IRL\cite{ng2000algorithms}. DPO maximizes the log-probability difference between preferred and rejected actions through entropy-regularized optimization.This theoretical analysis leads to the following important corollary:

\begin{corollary}
\label{corollary:two_stage}
MCTS search process in MCTS-EP implicitly performs cost function learning, while the DPO optimization corresponds to policy improvement with respect to this learned cost function. This two-stage process ensures both explorations through MCTS and effective policy updates through DPO.
\end{corollary}

\section{Experiments}
In this section, we conduct serval experiments to validate the effectiveness of MCTS-EP. Our framework demonstrates better performance compared to baselines and SOTA methods. 
\subsection{Experiment Settings}

\textbf{Datasets} We evaluate MCTS-EP on the ALFWorld \cite{shridhar2020alfworld} and WebShop \cite{yao2022webshop} benchmark. ALFWorld extends the ALFRED \cite{shridhar2020alfred} dataset, useing the AI2-THOR \cite{kolve2017ai2} simulator to create a realistic household environment. It features a cross-modal framework for embodied household tasks, combining high-fidelity visual environments with text-based counterparts. WebShop is a web-based problem-solving benchmark that tests agents to navigate an e-commerce website to locate and purchase products given requests from clients

\noindent \textbf{Training Setup}
We utilize Llama-3.1-8B-Instruct \cite{dubey2024llama} as the base model for verbal environments and Qwen2-VL-7B-Instruct \cite{wang2024qwen2} for vision-based tasks. Llama-3.2-Vision-11B-Instruct was excluded from our experiments due to issues with generating repetitive outputs and triggering warnings related to safety guideline violations. Detailed experimental settings and hyperparameters can be found in \cref{Experiment Settings}.

\noindent \textbf{Baselines}
To evaluate the effectiveness of our proposed method, we compare MCTS-EP against several baselines and state-of-the-art (SOTA) agents on the ALFWorld benchmark. The categorization and selection of baselines follow the methodology from EMMA \cite{yang2024embodied}. For textual environments, we evaluate three types of baselines: transformer-based models, such as BUTLER \cite{shridhar2020alfworld} and GPT-BUTLER \cite{micheli2021language}; feedback-driven methods, including ReAct \cite{yao2022react}, Reflexion \cite{shinn2024reflexion}, and DEPS \cite{wang2023describe}, which leverage reasoning traces, environmental feedback, or self-explanations to refine actions; and multi-agent collaboration approaches, such as AutoGen \cite{wu2023autogen}, which enable coordination between multiple LLM agents to tackle complex tasks. For visual environments, we evaluate several baselines, including MiniGPT-4 \cite{zhu2023minigpt}, BLIP-2 \cite{li2023blip}, LLaMA-Adaptor \cite{gao2023llama}, and InstructBLIP \cite{dai2023instructblip}, which integrate visual and textual understanding through VLM, as well as EMMA \cite{yang2024embodied}, a multi-modal agent designed specifically for embodied tasks with enhanced cross-modal reasoning capabilities. 
We adopt the baseline training methods from IPR \cite{xiong2024watch}, including PPO \cite{schulman2017proximal}, RFT \cite{yuan2023scaling}, and ETO \cite{song2403trial}, and additionally incorporate agent method like ReACT and Reflexion as WebShop baselines.

\begin{table*}[!h]
\caption{ALFWorld Success Rate}
\label{success_rate}
\vskip 0.1in
\begin{center}
\setlength{\tabcolsep}{1.5pt}
{\fontsize{8}{9}\selectfont
\begin{sc}
\begin{tabular}{@{}lcccccccc@{}}
\toprule
Type & Method & Success Rate & Pick\&Place & Clean\&Place & Heat\&Place & Cool\&Place & Look-in-Light & Pick-2\&Place \\ 
\midrule
Text & BUTLER & 0.26 & 0.31 & 0.41 & 0.69 & 0.27 & 0.12 & 0.29 \\
Text & GPT-BUTLER & 0.69 & 0.62 & 0.81 & 0.85 & 0.78 & 0.50 & 0.47 \\
Text & ReAct & 0.54 & 0.71 & 0.65 & 0.62 & 0.44  & 0.28 & 0.35 \\ 
Text & Reflexion & 0.91 & 0.96 & 1.00 & 0.81 & 0.83 & 0.94 & \textbf{0.88} \\
Text & DEPS & 0.76 & 0.93 & 0.50 & 0.80 & \textbf{1.00} & 1.00 & 0.00 \\
Text & AutoGen & 0.77 & 0.92 & 0.74 & 0.78 & 0.86  & 0.83 & 0.41 \\
Text & IPR & 0.75 & -- & -- & -- & -- & -- & -- \\
\midrule
Text & MCTS-EP (Ours) & \textbf{0.92} & \textbf{1.00} & \textbf{1.00} & \textbf{0.91} & 0.80 & \textbf{1.00} & 0.67 \\
\midrule
Vision & MiniGPT-4 & 0.16 & 0.04 & 0.00 & 0.19 & 0.17 & 0.67 & 0.06 \\
Vision & BLIP-2 & 0.04 & 0.00 & 0.06 & 0.04 & 0.11 & 0.06 & 0.00 \\
Vision & LLaMA-Adapter & 0.13 & 0.17 & 0.10 & 0.27 & 0.22 & 0.00 & 0.00 \\
Vision & InstructBLIP & 0.22 & 0.50 & 0.26 & 0.23 & 0.06 & 0.17 & 0.00 \\
Vision & EMMA & 0.82 & 0.71 & 0.94 & \textbf{0.85} & \textbf{0.83} & 0.88 & \textbf{0.67} \\
\midrule
Vision & MCTS-EP (Ours) & \textbf{0.87} & \textbf{0.90} & \textbf{1.00}  & 0.82  & 0.80  & \textbf{1.00} & 0.50 \\
\bottomrule
\end{tabular}
\end{sc}
}
\end{center}
\vskip -0.1in
\end{table*}

\noindent \textbf{Evaluation Metrics}
We evaluate all methods using standard metrics across different benchmarks. For the ALFWorld benchmark, we measure both the success rate (percentage of successfully completed tasks) and the average number of interaction steps required per task completion, which reflects the agent's decision-making efficiency. For the WebShop benchmark, following the evaluation protocol in IPR \cite{xiong2024watch}, we report the average reward computed over 200 test episodes.

\begin{table*}
\caption{ALFWorld Steps Counts}
\label{steps}
\vskip 0.1in
\begin{center}
\setlength{\tabcolsep}{1.5pt}
{\fontsize{8}{9}\selectfont
\begin{sc}
\begin{tabular}{@{}lcccccccc@{}}
\toprule
Type & Method & Steps & Pick\&Place & Clean\&Place & Heat\&Place & Cool\&Place & Look-in-Light & Pick-2\&Place \\ 
\midrule
Text & GPT-BUTLER & 18.8 & 18.4 & 18.4 & 12.7 & 15.6 & 24.4 & 26.6 \\
Text & ReAct & 20.6 & 18.1 & 18.8 & 18.2 & 23.2  & 23.7 & 25.5 \\ 
Text & Reflexion & 18.7 & 17.4 & 17.0 & 19.4 & 21.6 & 16.9 & 21.6 \\
\midrule
Text & MCTS-EP (Ours) & \textbf{10.2} & \textbf{11.5} & \textbf{9.2} & \textbf{12.3} & \textbf{8.8} & \textbf{8.7} & \textbf{11.8} \\
\midrule
Vision & MiniGPT-4 & 26.9 & 29.0 & 30.0 & 26.3 & 26.7 & 17.7 & 28.9 \\
Vision & BLIP-2 & 29.5 & 30.0 & 29.3 & 29.9 & 28.2 & 29.2 & 30.0 \\
Vision & LLaMA-Adapter & 27.5 & 26.4 & 28.6 & 24.2 & 26.7 & 30.0 & 30.0 \\
Vision & InstructBLIP & 26.2 & 21.5 & 25.0 & 27.2 & 28.9 & 26.8 & 30.0 \\
Vision & EMMA & 19.5 & 19.3 & 17.5 & 19.6 & 19.9 & 19.6 & 22.4 \\
\midrule
Vision & MCTS-EP (Ours) & \textbf{9.9} & \textbf{6.8} & \textbf{10.2}  & \textbf{13.1}  & \textbf{10.9}  & \textbf{6.4} & \textbf{15.3} \\
\bottomrule
\end{tabular}
\end{sc}
}
\end{center}
\vskip -0.1in
\end{table*}

\subsection{Experimental Results}
\textbf{Execution Time for MCTS-EP.} 
While computational cost remains a primary concern for MCTS algorithms, our optimized implementation achieves practical efficiency. Using Llama-3.1-8B as the policy model with search parameters (max depth=10, width=3, simulations=3), our MCTS-EP completes each ALFWorld task in 4 minutes on dual A100 GPUs. This yields a full search tree containing 87 SFT trajectories and 28 preference pairs per episode.
\begin{table}
\centering
\caption{WebShop Average Reward}
\label{webshop_result}
\setlength{\tabcolsep}{4pt} 

\begin{tabular}{c|c|c}
\hline
\textbf{Type} & \textbf{Setting} & \textbf{WebShop}  \\
\hline
\multirow{5}{*}{Training} & Llama-2-7B + PPO & 0.64 \\
                             & Llama-2-7B + RFT & 0.63 \\
     & Llama-3-8B + SFT & 0.61 \\
                             & Llama-3-8B + ETO & 0.66 \\
                             & Llama-3-8B + IPR & 0.72 \\
\hline
\multirow{4}{*}{Agent} & ReACT\cite{yao2022react} & 0.66 \\
                              & Reflexion \cite{shinn2024reflexion} & 0.35  \\
                        & ReACT(GPT-4o) & 0.54 \\
                        & ReACT(Claude 3.7) & 0.43\\
\hline
\multirow{1}{*}{Ours} & MCTS-EP & 0.81 \\
\hline
\end{tabular}
\label{tab:model_performance}
\end{table}

\textbf{MCTS-EP achieves SOTA in both visual and textual environments.} We evaluate MCTS-EP against all baselines in both textual and visual environments on the ALFWorld and WebShop. Results are summarized in \cref{success_rate}, \cref{steps} and \cref{webshop_result} with baseline results sourced from \cite{yang2024embodied} and \cite{xiong2024watch}. Our proposed method, MCTS-EP, demonstrates significant improvements over all baselines, achieving state-of-the-art performance in both environments.

\textbf{Optimized Task Completion Rate.} MCTS-EP outperforms all baselines in ALFWorld success rate, surpassing the previous textual SOTA Reflexion (91\%) \cite{shinn2024reflexion}  and visual SOTA EMMA (82\%) \cite{yang2024embodied}. Notably, Reflexion and EMMA require multiple attempts to achieve their final results, whereas our method completes tasks in a single attempt. These performances gains can be primarily attributed to the online nature of MCTS-EP, which enables dynamic planning and decision-making during task execution. Online MCTS-EP is able to handle dynamic and partially observable environments effectively. This flexibility allows the agent to generalize its strategies and maintain high success rates across diverse tasks, regardless of their complexity or dependencies.
\begin{table*}
\caption{ALFWorld Ablation Study Result}
\label{tab:ablation}
\vskip 0.1in
\begin{center}
\setlength{\tabcolsep}{1.5pt}
{\fontsize{8}{9}\selectfont  
\begin{sc}

\begin{tabular}{@{}l|ccccccc@{}}
\toprule
Method & Success Rate & Pick\&Place & Clean\&Place & Heat\&Place & Cool\&Place & Look-in-Light & Pick-2\&Place \\ 
\midrule
w/o Fine-Tuning ($\mathcal{B}$) & 0.53 (8.7) & 0.60 (4.8) & 0.57 (9.5) & 0.36 (14.0) & 0.80 (9.9) & 0.56 (6.0) & 0.17 (9.0) \\
w/o DPO  & 0.87 (11.8) & 0.90 (10.3) & 0.93 (9.8) & 0.91 (16.9) & 0.80 (11.4) & 1.00 (9.1) & 0.50 (16.7) \\
MCTS-EP (Ours)  & 0.87 (9.9) & 0.90 (6.8) & 1.00 (10.2)  & 0.82 (13.1)  & 0.80 (10.9)  & 1.00 (6.4) & 0.50 (15.3) \\
\bottomrule
\end{tabular}
\end{sc}
}
\end{center}
\vskip -0.1in
\end{table*}

\begin{table}
\centering
\caption{WebShop Ablation Study Result}
\label{webshop_ablation}
\setlength{\tabcolsep}{4pt} 

\begin{tabular}{c|c}
\hline
\textbf{Method} & \textbf{Average Reward} \\
\hline
w/o Fine-Tuning ($\mathcal{B}$)& 0.61 \\
\hline
w/o DPO & 0.68 \\
\hline
MCTS-EP (Ours) & 0.81\\
\hline
\end{tabular}
\label{tab:model_performance}
\end{table}

\textbf{Fewer Steps, Better Task Performance.} MCTS-EP reduces the interaction steps to 10.22 in text and 9.9 in vision ALFWorld, a significant improvement compared to Reflexion’s 18 and EMMA's 19.5. 
The performance gains are primarily attributed to DPO, allowing the embodied agent to refine its commonsense understanding of the environment. This ensures that the agent focuses on the most relevant actions to achieve task objectives, avoiding unnecessary exploration. These results also emphasize the model's capability to integrate cross-modal reasoning seamlessly. By leveraging its multi-modal understanding, MCTS-EP achieves greater efficiency in environments requiring both textual and visual inputs.

\textbf{Optimization in High-Dimensional Action Space.}
MCTS-EP also achieves SOTA performance on the Webshop benchmark, which has a high-dimensional action space. While standard MCTS struggles with large action spaces, and Reflexion \cite{shinn2024reflexion} admits difficulty in tasks requiring extensive exploration, our method overcomes these limitations. 
By iteratively guiding the policy, we dynamically prune low-value actions—suppressing their probabilities—to focus exploration efficiently, achieving a final average reward of 0.81.

\subsection{Ablation Study}
To evaluate the impact of individual components in out framwork, we conducted ablation experiments on the two key steps of MCTS-EP: Fine-tuning with success trajectory data and DPO. The results are presented in Table \ref{tab:ablation} and Table \ref{webshop_ablation}.
In our table, "w/o Fine-Tuning ($\mathcal{B}$)" represents that we omit the MCTS exploration phase and fine-tune the model solely using expert trajectory data. "w/o DPO" indicates that we remove DPO training with preference pairs obtained from MCTS exploration, while retaining the fine-tuning process with successful trajectories discovered through MCTS. MCTS-EP is complete method.

\textbf{Online Fine-tuning Enhances Accuracy.} Removing the fine-tuning step with success trajectory results in a significant drop in performanace, with the success rate decreasing from 87\% to 53\%. This demonstrates the critical role of fine-tuning success trajectory data in improving the model's ability to adapt to task-specific requirements and align its outputs with desired goals.

\textbf{DPO Improves Commonsense.} When DPO is removed, the success rate remains at 87\%, the same as the full framework. However, the average interaction steps increase significantly, from 9.9 to 11.8, indicating a loss in decision-making efficiency. This highlights how DPO helps refine the model’s commonsense reasoning and action prioritization, enabling the agent to make more efficient decisions during task execution.

\section{Conclusion}
This paper presents MCTS-EP, an online framework for training embodied agents that combines MCTS with DPO. Our approach addresses key challenges in embodied environments based on three components: (1) an efficient preference data collection method, (2) a selective state representation strategy for multi-modal reasoning, and (3) an online framework that iteratively improves model performance.

We provide theoretical analysis for our framework, proving that MCTS-based policies achieve better performance bounds. MCTS-EP demonstrates SOTA performance in both textual and visual environments. Ablation studies highlight the roles of the framework's components: fine-tuning with success trajectories enhances task completion accuracy, while DPO significantly improves the model's commonsense reasoning. These results suggest that combining MCTS-guided exploration with preference optimization creates a robust foundation for training embodied agents.

\bibliography{example_paper}
\bibliographystyle{icml2025}

\newpage
\appendix
\onecolumn
\section{Proofs}
\subsection{Proof of Theorem \ref{theorem:Performance}}  
\label{app:Performance}
Each policy $\pi_i$ is a mixed policy combining the expert policy $\pi^*$ and learned policy $\hat{\pi}_i$ with a non-increasing mixing coefficient $\beta_i$, expressed as $\pi_i = \beta_i\pi^* + (1-\beta_i)\hat{\pi}_i$, where the sequence $\hat{\pi}_{1:N}$ represents the policies $\hat{\pi}_1$ through $\hat{\pi}_N$. The term $\epsilon_N$ defines the loss of the best policy in hindsight after N iterations, calculated as $\min\limits_{\pi \in \Pi}\frac{1}{N}\sum_{i=1}^N \mathbb{E}_{s \sim d_{\pi_i}}[\ell(s,\pi)]$. The variable $\ell_{max}$ serves as an upper bound on the loss, ensuring that $\ell_i(s,\hat{\pi}_i) \leq \ell_{max}$ holds for all policies $\hat{\pi}_i$ and states $s$ where $d_{\hat{\pi}_i}(s) > 0$. The term $n_\beta$ represents the largest value of n not exceeding N that satisfies $\beta_n > \frac{1}{T}$, while $\gamma_N$ denotes the average regret observed across the policy sequence $\hat{\pi}_{1:N}$.

\begin{lemma}\label{lemma:dagger}
For DAgger, there exists a policy $\hat{\pi} \in \hat{\pi}_{1:N}$ such that:
$$\mathbb{E}_{s \sim d_{\hat{\pi}}}[\ell(s, \hat{\pi})] \leq \epsilon_N + \gamma_N + \frac{2\ell_{max}}{N}[n_\beta + T\sum_{i=n_\beta+1}^N \beta_i]$$
\end{lemma}
Based on Lemma \ref{lemma:dagger}, we can prove that MCTS has better performance bounds than on-policy algorithms.

\begin{proof}

We begin by expressing the PUCT policy in terms of UCT and prior policy $\pi_{\theta}$. Define policy model $\pi_{uct} = Q(s,a) + c\frac{\sqrt{\sum_b n(x,b)}}{1 + n(x,a)}$, where for each step $i$, $Q(s_i,a)$ will update, so $\pi_{uct}$ can be interpreted as iterative policy model $\pi_{uct}^i$ with appropriate scaling.

The MCTS with PUCT policy can be written as:
$$
\pi_{mcts}^i = \pi_{uct}^i + c *\frac{\sqrt{\sum_b n(x,b)}}{1 + n(x,a)} (\pi_{\theta} - 1).
$$

Let $c_i = c\frac{\sqrt{\sum_bn(x,b)}}{1 + n(x,a)}$, we can rewrite $\pi_{mcts}$ as:

$$
\pi_{mcts}^i \propto \frac{1}{c_i + 1} \pi_{uct}^i + \frac{c_i}{1 + c_i}(\pi_{\theta}-1)
= (1 - \beta_i) \pi_{uct}^i + \beta_i (\pi_{\theta} -1 )
$$
where $\beta_i = \frac{c_i}{1 + c_i}$. Note that the $\propto$ holds because we use $\argmax$ for action selection in $\pi_{mcts}$, and scaling by a constant $(1 + c_i)$ does not affect the $\argmax$ operation.

The proof relies on three key conditions:

1. UCT Regret Bound: From \cite{auer2002finite}, the cumulative regret $R(T)$ of UCT  grows as $O(\ln n)$ in specific cases, it is a no-regret algorithm:
   $$\lim_{T \to \infty} \frac{R(T)}{T} = 0$$

2. Decay of Mixing Coefficient: As $N \to \infty$, $\frac{c_i}{1 + c_i}$ decreases to a positive constant. To ensure $\overline{\beta_N} = \frac{1}{N}\sum_i^N \beta_i \to 0$ as $N \to \infty$, we can simple set $\beta_i = I(i=1)$.

3. Loss Function Property: The loss function is strongly convex.

Let $\epsilon_N = min_{\pi \in \mathcal{\pi}} \frac{1}{N}\sum_{i=1}^N \mathbb{E}_{s \sim d_{\pi_{mcts}^i}}[l(s,\pi)]$ represent the loss of the best policy.
From Lemma \ref{lemma:dagger}, there exists a policy $\pi_{uct}^i$ such that:
\begin{equation}
\mathbb{E}_{s \sim d_{{\pi}_{uct}}} (l(s,\pi_{uct})) \leq \epsilon_N + \gamma_N + \frac{2l_{max}}{N}[n_{\beta}+T\sum_{i=n_{\beta}+1}^N \beta_i],
\end{equation}
where $\gamma_N$ is the average regret of $\pi_{uct}^0$ to $\pi_{uct}^N$. As $N \to \infty$, $\pi_{uct}$ is no-regret, then $\gamma_N \to 0$ and the last term approaches 0.

Since $\epsilon_N$ represents the loss of the optimal policy, we have $\epsilon_N \leq \mathbb{E}_{s \sim d_{\pi_{mcts}}}[l(s, \pi_{\theta})]$. Furthermore, given that the loss function $l(s, \cdot)$ is convex in $\pi$ for all states $s$, we can derive:
\begin{equation}
\begin{aligned}
\mathbb{E}_{s \sim d_{\pi_{mcts}}} [l(s,\pi_{mcts})] &\leq (1 - \beta_i) \mathbb{E}_{s \sim d_{\hat{\pi_{i}}}} [l(s,\hat{\pi_{i}})] + \beta_i \mathbb{E}_{s \sim d_{\pi_{mcts}}}[l(s, \pi_{\theta})] \
&\leq \mathbb{E}_{s \sim d_{\pi_{mcts}}}[l(s, \pi_{\theta})]
\end{aligned}
\end{equation}

\end{proof}

\subsection{Proof of Theorem \ref{theorem:gail}}  
\label{app:gail}
\begin{proof}
Inverse reinforcement learning (IRL) \cite{ng2000algorithms} primitive procedure can be defined as 
\begin{equation}
\begin{aligned}
\label{eq:IRL}
IRL_{\psi}(\pi_E) = \underset{c \in \mathbb{R}^{\mathcal{S} \times \mathcal{A}}}{\argmax}
-\psi(c) + 
\left( \min_{\pi \in \Pi} -H(\pi) + \mathbb{E}_{\pi}[c(s, a)] \right) 
- \mathbb{E}_{\pi_E}[c(s, a)]
\end{aligned}
\end{equation}
where $\psi$ is cost regularization, $c$ is the cost function, $H(\pi) = \mathbb{E}_{\pi}[-\log \pi(a|s)]$, $\pi$ is learned policy and $\pi_E$ is expert policy. It is a max-min problem. IRL can generate optimal cost function.
Define $RL(c)$ to generate optimal policy:
\begin{equation}
RL(c) = \underset{\pi \in \Pi}{\argmin} \, -H(\pi) + \mathbb{E}_{\pi}[c(s, a)]
\end{equation}
If $\tilde{c} \in \text{IRL}_{\psi}(\pi_E)
$, we want to get optimal policy by $RL(\tilde{c})$. Then the dual problem of equation \ref{eq:IRL} with some relax is :
\begin{equation}
\text{RL} \circ \text{IRL}_{\psi}(\pi_E) =\underset{\pi \in \Pi}{\argmin} \, -H(\pi) + \psi^*(\rho_{\pi} - \rho_{\pi_E}).
\end{equation}
where $\rho_{\pi}(s, a) = \pi(a \mid s) \sum_{t=0}^{\infty} \gamma^t P(s_t = s \mid \pi)$ can be interpreted as the distribution of state-action pairs that an agent encounters when navigating the environment with policy $\pi$, $\psi^*$ is the dual function of cost regularization $\psi$.

The proof consists of three key steps:

1) First, we establish that $\psi(a)$ is a valid convex regularization function on $S \times A \rightarrow \mathbb{R}$. Its convex conjugate is:
\begin{align}
\label{convex conjugate}
\psi^*(\rho_{\pi} - \rho_{\pi_E}) &= \sup_a \{(\rho_{\pi} - \rho_{\pi_E})^Ta- \psi(a)\} \\
&= -\mathbb{E}_{\pi_E}[\log \pi_{E}(a|s)]+ \sup_a \{(\rho_{\pi} - \rho_{\pi_E})^Ta - c*\frac{\sqrt{ln t}}{t*\rho_{\pi}(s,a)}\}.
\end{align}

2) The occupancy measure $\rho_{\pi}(s,a) = \pi(a|s)\sum_{t}\gamma^{t}P(s_t = s| \pi)$ represents the discounted visitation frequency of state-action pairs under policy $\pi$. In the MCTS context, $t*\rho_{\pi}(s,a)$ is proportional to the visit count $N(s,a)$, as both measure the frequency of state-action selections.

3) The second term in equation \ref{convex conjugate} corresponds to the Lower Confidence Bound (LCB) formulation approximately \cite{rashidinejad2021bridging}. By defining $\pi_E$ as the policy derived from preferred actions and $\pi$ as the policy over rejected actions, we obtain:

\begin{equation}
\begin{aligned}
\text{RL} \circ \text{IRL}_{\psi}(\pi_E) = & \underset{\pi \in \Pi}{\argmin} -\mathbb{E}_{\pi_E}[\log \pi_{E}(a|s)] + \mathbb{E}_{\pi}[\log \pi(a|s)] \\
& + \sup_a \{(\rho_{\pi} - \rho_{\pi_E})^Ta - c*\frac{\sqrt{ln t}}{t*\rho_{\pi}(s,a)}\}
\end{aligned}
\end{equation}

This formulation reveals a two-stage optimization process:
\begin{itemize}
    \item Stage 1 (IRL): The LCB policy generates a value function serving as the cost function in IRL
    \item Stage 2 (RL): DPO maximizes the log-probability difference between preferred and rejected actions
\end{itemize}

This completes the connection between MCTS-EP and the GAIL framework.
\end{proof}

\section{Experiment Settings}
\label{Experiment Settings}
\subsection{Dataset Details}
ALFWorld tasks involve natural language goal descriptions (e.g., “put a pan on the dining table”) and require agents to perform high-level actions such as go to, clean and opening. The framework includes six task types: Pick \& Place, Clean \& Place, Heat \& Place, Cool \& Place, Look in Light, and Pick Two Objects \& Place. These tasks challenge agents to navigate visually diverse environments, interact with up to 120 dynamically populated rooms, and manipulate portable objects (e.g., apples, mugs) and static receptacles (e.g., drawers, microwaves). To enable cross-modal learning, ALFWorld uses the TextWorld engine \cite{cote2019textworld} to generate text-based analogs of ALFRED scenes, described using the Planning Domain Definition Language (PDDL) \cite{aeronautiques1998pddl}. Each scene’s state and dynamics are mirrored across modalities, allowing agents to train and evaluate in either the visual or text-based environment.
\subsection{Hyperarameters Details}

For single-modal training, we apply LoRA \cite{hu2021lora} fine-tuning on Llama-3.1-8B-Instruct with a sequence length of 2048, a device batch size of 1, and a learning rate of 5e-5 using a cosine scheduler, fine-tuning for 2 epochs with 8 gradient accumulation steps. For multi-modal tasks, we use Qwen2-VL-7B-Instruct, unfreezing the vision tower during fine-tuning with a learning rate of 1e-4 and a batch size of 4 per device. For multi-modal DPO, only the backbone is fine-tuned, with the learning rate reduced to 5e-6 for stability. Across all experiments, we set the LoRA rank to 16 and use a sigmoid preference loss function with $\beta$ set to 0.5.

\subsection{System Prompts}

\begin{figure}[h!]
\centering
\begin{tcolorbox}[colback=white, colframe=black, title=Text-Only Prompt]
You are an AI robot agent in an interactive environment. Your goal is to accomplish the given task through a series of actions. Follow these guidelines:
\begin{itemize}
    \item Carefully analyze the task requirements. Break down complex tasks into smaller, manageable steps and create a mental plan before acting
    \item Be persistent in searching for required objects. When searching for objects, use common sense to predict likely locations, then systematically explore those areas
    \item For tasks involving multiple objects, keep a mental count of how many you've collected or placed
    \item Avoid repeating the same action consecutively. If an action doesn't work, explore other objects or locations
    \item Your response must be exactly one action name chosen strictly from provided candidate actions
\end{itemize} 
\end{tcolorbox}
\caption{Text-Only Prompt}
\end{figure}

\clearpage
\vspace*{0pt}

\begin{figure}[h!]
\centering
\begin{tcolorbox}[colback=white, colframe=black, title=Multi-Modal Prompt]
You are an AI robot agent in an interactive environment. Your goal is to accomplish the given task through a series of actions. Follow these guidelines:
\begin{itemize}
    \item Carefully analyze the task requirements. When searching for objects, use common sense to predict likely locations
    \item For tasks involving multiple objects, keep a mental count of collected or placed items
    \item Respond only with a JSON object containing:
    \begin{itemize}
        \item \textbf{"what\_you\_see"}: your detailed observation
        \item \textbf{"action"}: one action name
    \end{itemize}
\end{itemize}
\end{tcolorbox}
\caption{Multi-Modal Prompt}
\end{figure}

\end{document}